\documentclass[runningheads]{llncs}
\usepackage[bookmarks=true, pdftex, colorlinks=true, plainpages=false,
pdfauthor={}, pdftitle={}, pdfkeywords={}]{hyperref}
\usepackage{graphicx}
\usepackage{amsfonts}
\usepackage{color}
\DeclareGraphicsExtensions{.jpg,.pdf,.mps,.png,.ps}
\usepackage{booktabs}
\usepackage{microtype}
\usepackage{rotating}
\usepackage{bm}
\usepackage{cite}

\newcommand{\hn}{\phantom{-}}

\begin{document}
\pdfpageattr {/Group << /S /Transparency /I true /CS /DeviceRGB>>}

\title{Topology guaranteed segmentation of the human retina from OCT using convolutional neural networks}

\titlerunning{Topology guaranteed segmentation of human retina OCT using CNNs}

\author{Yufan He\inst{1} \and Aaron Carass\inst{1,2} \and Bruno M.
Jedynak\inst{3} \and Sharon~D.~Solomon\inst{4} \and Shiv
Saidha\inst{5} \and Peter A. Calabresi\inst{5} \and
Jerry~L.~Prince\inst{1,2}}

\authorrunning{Y.~He et al.}

\index{He, Yufan}
\index{Carass, Aaron}
\index{Jedynak, Bruno M.}
\index{Solomon, Sharon D.}
\index{Saidha, Shiv}
\index{Calabresi, Peter A.}
\index{Prince, Jerry L.}

\institute{$^{1}$Dept. of Electrical and Computer Engineering,
$^{2}$Dept. of Computer Science,\\
The Johns Hopkins University, Baltimore, MD~21218,~USA\\[0.75em]
$^{3}$Dept. of Mathematics \& Statistics, Portland State
University,\\
Portland, OR 97201, USA\\[0.75em]
$^{4}$Wilmer Eye Institute, $^{5}$Dept. of Neurology,\\
The Johns Hopkins University School of Medicine, MD~21287,~USA
}


\maketitle

\begin{abstract}
Optical coherence tomography~(OCT) is a noninvasive imaging modality
which can be used to obtain depth images of the retina. The changing
layer thicknesses can thus be quantified by analyzing these OCT
images, moreover these changes have been shown to correlate with
disease progression in multiple sclerosis. Recent automated retinal
layer segmentation tools use machine learning methods to perform
pixel-wise labeling and graph methods to guarantee the layer hierarchy
or topology. However, graph parameters like distance and smoothness
constraints must be experimentally assigned by retinal region and
pathology, thus degrading the flexibility and time efficiency of the
whole framework. In this paper, we develop cascaded deep networks to
provide a topologically correct segmentation of the retinal layers in a
single feed forward propagation. The first network~(S-Net) performs
pixel-wise labeling and the second regression network~(R-Net) takes
the topologically unconstrained S-Net results and outputs layer
thicknesses for each layer and each position. Relu activation is used
as the final operation of the R-Net which guarantees non-negativity of the output layer thickness. Since the segmentation
boundary position is acquired by summing up the corresponding
non-negative layer thicknesses, the layer ordering~(i.e., topology) of the
reconstructed boundaries is guaranteed even at the fovea where the
distances between boundaries can be zero. The R-Net is trained using
simulated masks and thus can be generalized to provide topology
guaranteed segmentation for other layered structures. This deep
network has achieved comparable mean absolute boundary
error~(2.82~$\mu$m) to state-of-the-art graph methods~(2.83~$\mu$m).
\end{abstract}
\keywords{Retina OCT, Deep learning segmentation, Topology guarantee.}

\section{Introduction}
\label{s:intro}
Optical coherence tomography~(OCT) is a widely used non-invasive and
non-ionizing modality for retina imaging which can obtain 3D retina
images rapidly~\cite{hee1995archo}. The depth information of the
retina from OCT enables measurements of layer thicknesses, which are
known to change with certain diseases~\cite{medeiros2009iovs}. Fast
automated retinal layer segmentation tools are crucial for large
cohort studies of these diseases.\par
Automated methods for retinal layer segmentation have been well
explored~(\cite{rathke2014mia, antony2014miccai}).
State-of-the-art methods use machine learning (e.g,~random
forest~(RF)~\cite{lang2013boe}) for coarse pixel-wise labeling and
then level set~\cite{carass2014boe} or graph
methods~\cite{garvin2009tmi, lang2013boe} to guarantee the
segmentation topology~(i.e., the anatomically correct retinal layer ordering) and obtain the final boundary surfaces. They are
limited by the manually selected features for the pixel-wise labeling
task and the manually tuned parameters of the graph. To build the
graph, boundary distances and smoothness constraints which are
spatially varying need to be experimentally assigned. The manually
selected features and fine tuned graph parameters limit the
application across cohorts.\par
Deep learning automatically extracts relevant image features from
the training data and performs the segmentation in a feed forward
fashion. The fully convolutional network~(FCN) proposed by Long et
al.~\cite{long2015cvpr} is a successful deep learning segmentation
method and the U-Net variant~\cite{ronneberger2015miccai} is widely
used for medical image segmentation. Both Roy et
al.~\cite{roy2017arxiv} and He et al.~\cite{he2017towards} proposed
FCNs for retinal layer segmentation~(the former also included fluid
segmentation). However, these FCN methods provide pixel-wise
labeling without explicitly utilizing high level priors like shape,
and neither guarantee the correct topology. Examples of FCNs giving
anatomical infeasible results are shown in Fig.~\ref{UN_DNS}.\par
In order to obtain structured output directly from deep networks,
Zheng et al.~\cite{Zheng_2015_ICCV} implemented conditional random
field as a recurrent neural network. This method can provide better
label consistency but cannot guarantee global topology. BenTaieb et
al.~\cite{bentaieb2016miccai} proposed to explicitly integrate the
topology priors into the loss function during training and Romero et
al.~\cite{romero2017image} used a second auto-encoder network to learn
the output shape prior. Although those methods can improve the
segmentation results by utilizing shape and topology priors, they
still cannot guarantee the correct topology.\par
%
%
%
To obtain a topologically correct segmentation of the retinal layers
from a deep network in a single feed forward propagation, we propose
a cascaded FCN framework that transforms the layer segmentation
problem from pixel labeling into a boundary position regression
problem. Instead of outputting the boundary position directly, we use
the network to output the distance between two boundaries, i.e,
the layer thickness. The first network~(S-Net) performs pixel
labeling and the second regression network~(R-Net) takes the
topologically unconstrained S-Net results and outputs layer
thicknesses for each layer and each position.
Relu~\cite{dahl2013improving} activation is used as the final
operation of R-Net, which guarantees the non-negativity of the output
layer thicknesses. Since the boundary position is acquired
by summing up the corresponding non-negative layer thicknesses, the
ordering of the reconstructed boundaries is guaranteed even at the
fovea where the distances between boundaries can be zero.

\section{Method}
\begin{figure}[!tb]
\centering
\includegraphics[width =1\textwidth, clip, trim=0cm 0.5cm 0cm 0cm]{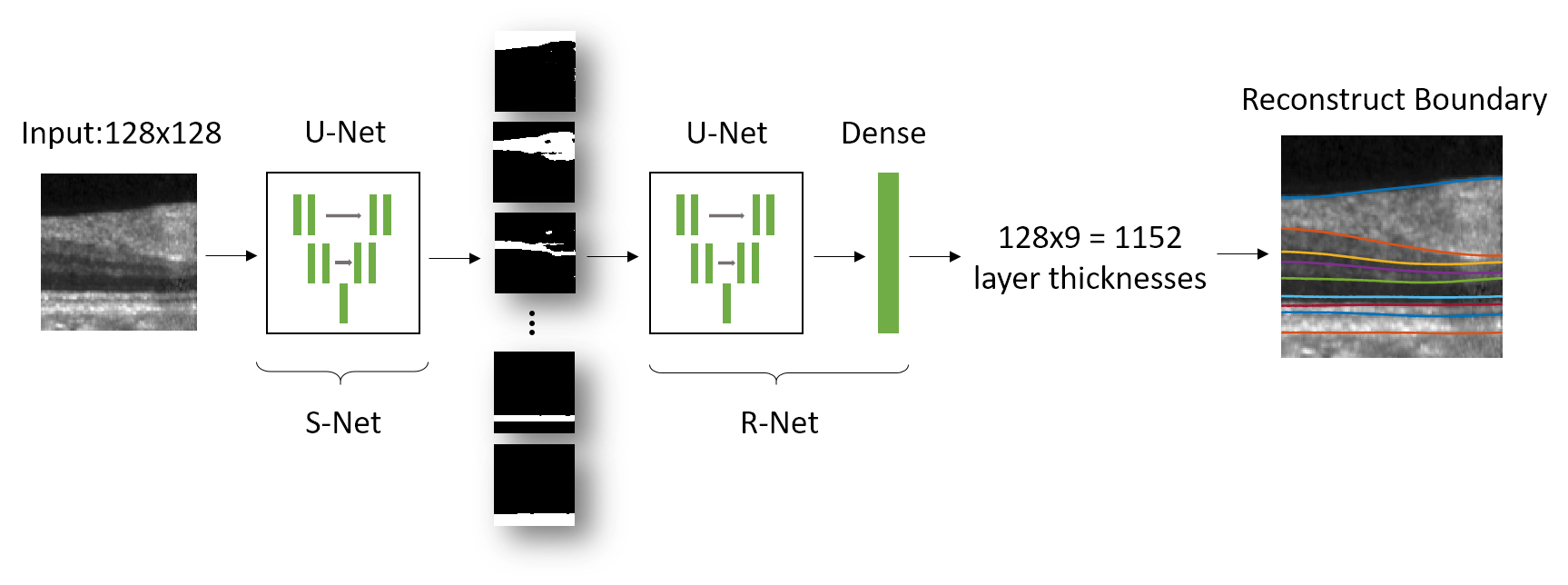}
\caption{A schematic of the proposed method.}
\label{framework}
\end{figure}

Fig.~\ref{framework} shows a schematic of our framework. We describe
each step in our processing below.

\noindent\textbf{Preprocessing}
A typical retinal B-scan is $496\times 1024$ which can require large
amounts of memory if directly processed by a deep network. To address
this, we approximately identify the Bruch's membrane, flatten the
retina and crop the image to remove the black background. Overlapped
patches of size $128\times128$ are extracted and segmented by the deep
network.
%
%
%

\begin{figure}[!b]
\centering
\includegraphics[width =1\textwidth, clip, trim=0cm 2cm 0cm 0cm]{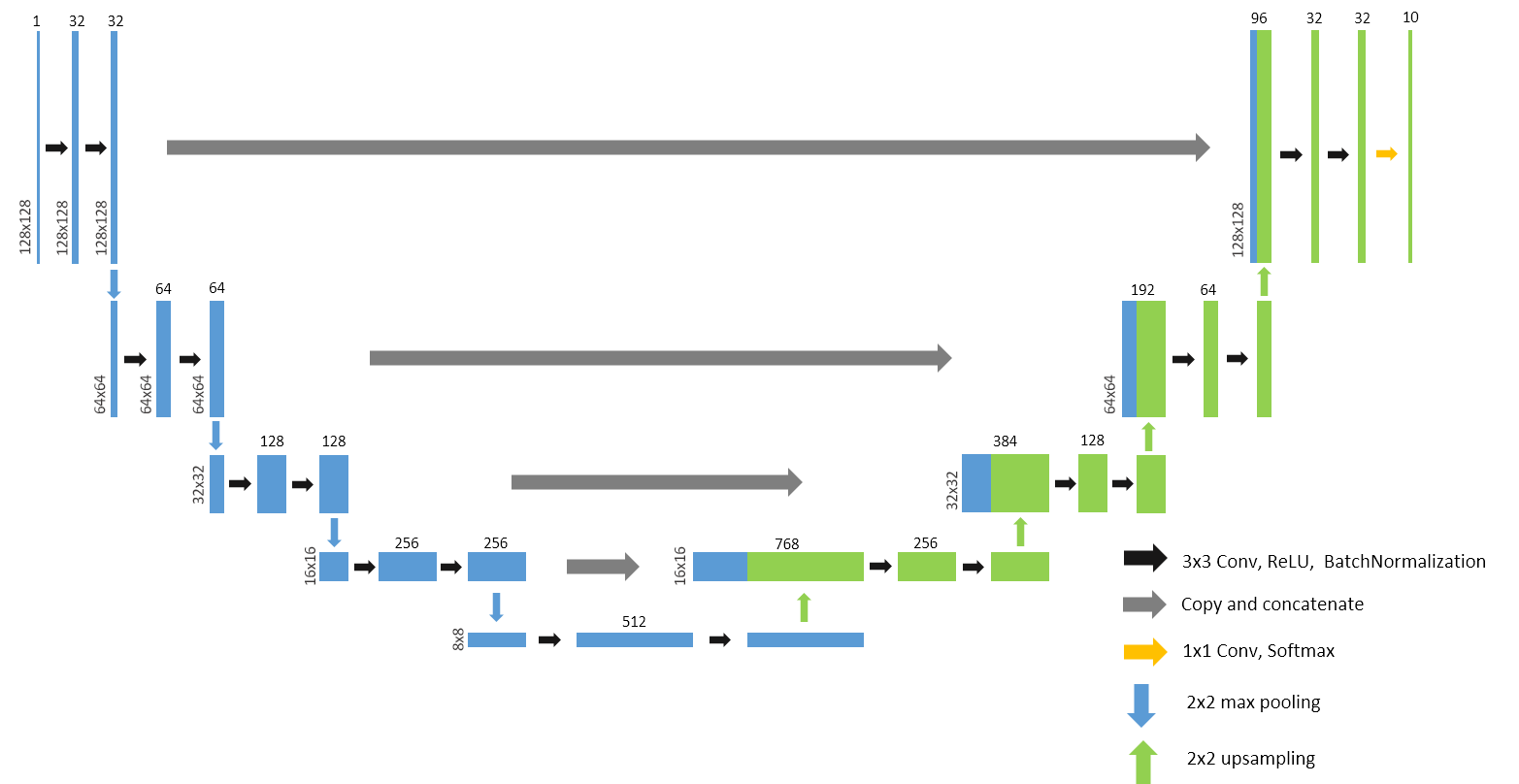}
\caption{A schematic of the of S-Net.}
\label{unet}
\end{figure}

\noindent\textbf{Segmentation Network~(S-Net) Overview}
Our segmentation FCN~(S-Net) is based on the 
U-Net~\cite{ronneberger2015miccai}. It takes a $128\times 128$ image
as input and the output is a $10 \times 128\times 128$
segmentation probability map which includes probability maps
for the eight retinal layers and the background above and below the
retina~(vitreous and choroid, respectively). Fig.~\ref{unet} shows
the details of S-Net; specifically, four $2\times 2$ max pooling and 19
convolution layers are used.

\noindent\textbf{Regression Net~(R-Net) Overview}
\noindent The R-Net consists of two parts: a U-Net identical to our
S-Net (except for the input channels) and a dense layer. The input to the
R-Net are the topologically unconstrained results from the S-Net.
R-Net is applied to learn the shape and topology priors of the layer
structures, while being resistant to the segmentation defects.~(see
Fig.~\ref{UN_DNS} for examples). The dense layer of the R-Net uses Relu activation and thus guarantees a non-negative output vector. The size of this output is $128 \times 9$ =
1152, corresponding to the thicknesses of the 9 layers over the 128
A-scans being segmented.

\subsection{Training}
We train our framework in two steps: S-Net is trained with a common
pixel-wise labeling scheme because every pixel in the training data
can be treated as an independent training sample and the total
training data size is enlarged~\cite{dou20173d}. R-Net is trained with
augmented ground truth masks to learn the shape and topology prior. An
alternative way to train the R-Net is to take the S-Net output as
input and output the ground truth layer thickness. However, training in
this manner would be sub-optimal as the S-Net output is not the
ground truth mask. Thus, the training pairs of the S-Net output and the
ground truth thicknesses are biased, which would bias the resultant
R-Net. Therefore, we train both networks independently. We note that
training the R-Net separately with simulated training masks allows
this network to be generalized for use with other layered structures.

\noindent\textbf{S-Net training}
\label{s-train}
The S-Net is trained with a common pixel-wise labeling scheme, namely
the cross-entropy loss function:
\begin{equation}
\mathcal{L} = -\sum\limits_{x\in \Omega}
g_l(x)\textrm{log}(p_l(x;\theta)).
\end{equation}
Here, $g_l(x)$ is an indicator function on the ground truth label of
pixel $x$ and $p_l(x;\theta)$ is the prediction probability from the deep
network that the pixel $x$ belongs to layer $l$. Standard
back-propagation is used to minimize the loss and update the network
parameter $\theta$.

\begin{figure}[!b]
\centering
\includegraphics[width =1\textwidth]{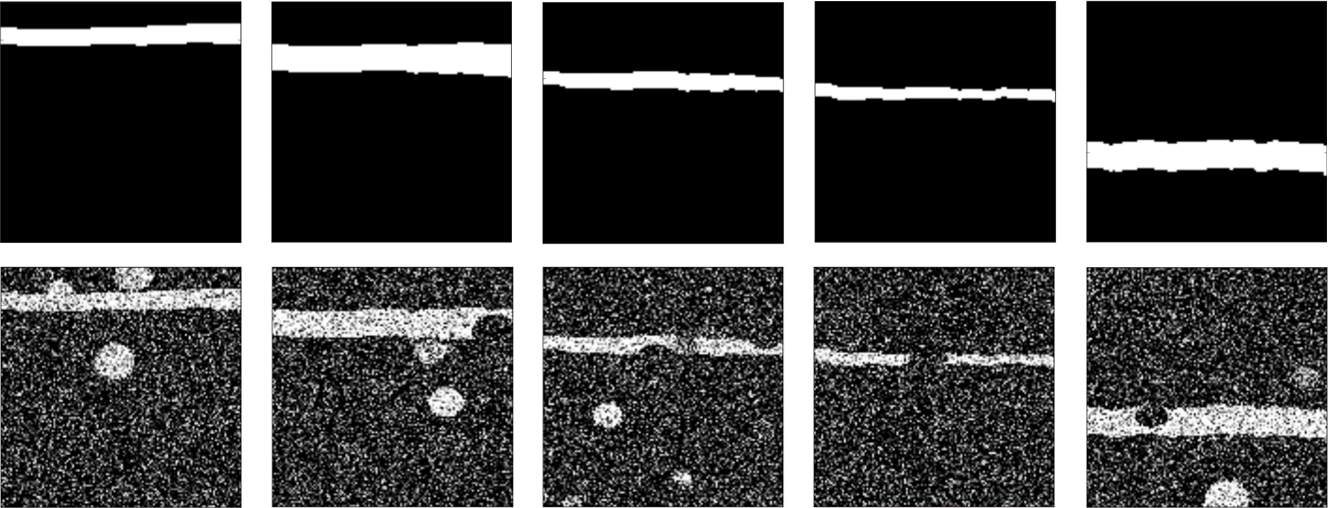}
\caption{Top row: Five $g_l(x)$ masks. Bottom row: After adding
noise and defects.}
\label{simulate}
\end{figure}

\noindent\textbf{R-Net training}
\label{r-train}
The purpose of the regression net is to find a mapping from the
pixel-wise segmentation probability maps into layer thicknesses. We
simulate topology defects with the ground truth mask and use R-Net to
recover the correct layer thicknesses. The training of R-Net is based
on minimizing the mean squared loss function below with standard
back-propagation,
\begin{equation}
\mathcal{L} = ||T(g(x)) - \mathcal{R}(g(x)+s(x);\theta)||_2^2.
\end{equation}
Here, $g(x)$ is the ground truth mask, $T(g(x))$ is the corresponding
ground truth layer thickness, $\mathcal{R}$ is the prediction from the
regression net, $s(x)$ is the simulated defects and Gaussian
noise~\cite{romero2017image} added to the ground truth mask. The
simulated defects are random ellipses with magnitude ranging from $-1$
to $1$. Examples of the simulated input masks to R-net are shown in
Fig.~\ref{simulate}.
%
%
To prevent R-net from over-fitting the thickness values, we randomly
move the position of the masks vertically and dilate or shrink the
masks.

\section{Experiments}
Ten fully delineated Spectralis Spectral Domain OCT (SD-OCT) scans~(of
size $496\times 1024\times 49$) were used for training. 20 overlapped
patches were extracted within each B-Scan for training both networks,
which yielded $9600$ samples for training. 20 SD-OCT
macular OCT scans~(of size $496\times 1024\times 49$) were acquired
for validation. Ten data sets in our validation cohort were diagnosed
with multiple sclerosis~(MS) and the remaining ten were healthy
controls.

\subsection{Results}
\begin{figure}[tb!]
\centering
\begin{tabular}{cc}
\includegraphics[width =0.5\textwidth]{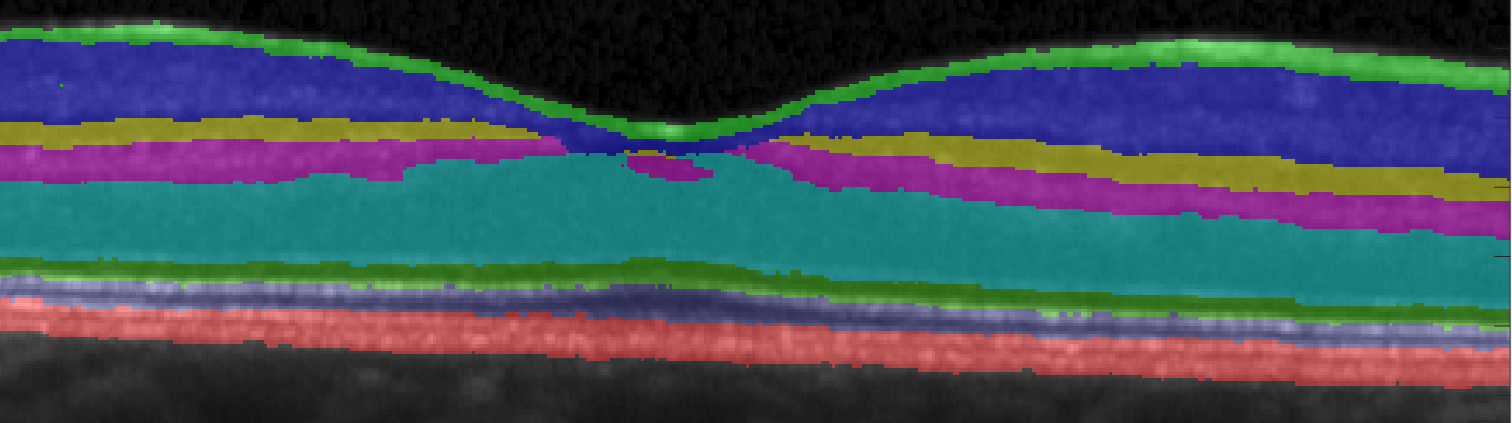}&
\includegraphics[width =0.5\textwidth]{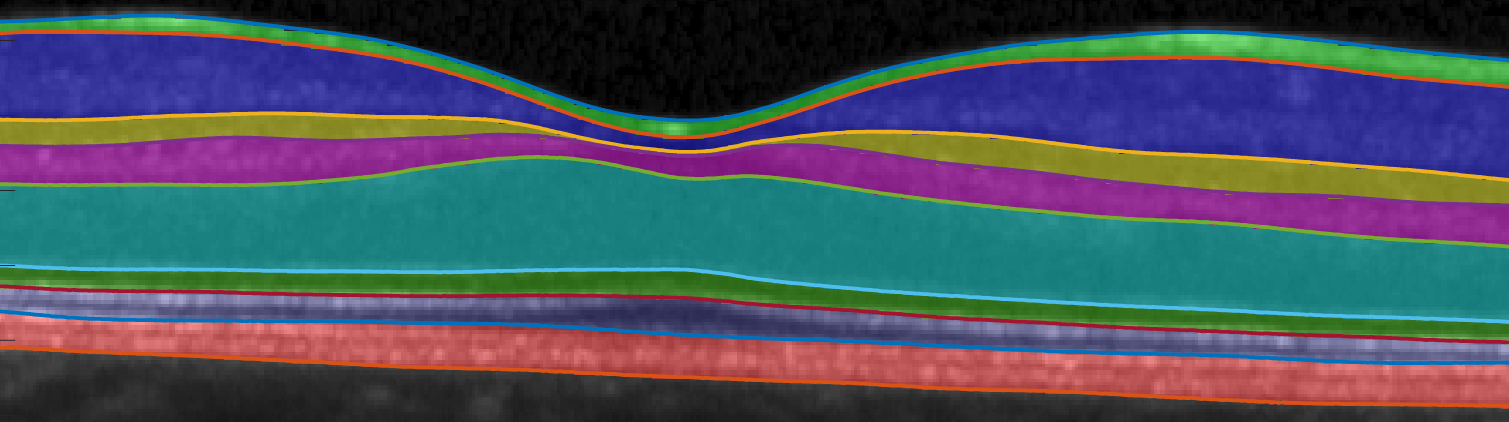}\\
\includegraphics[width =0.5\textwidth]{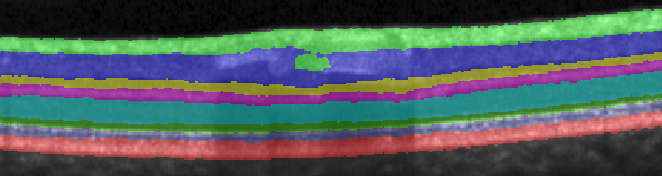}&
\includegraphics[width =0.5\textwidth]{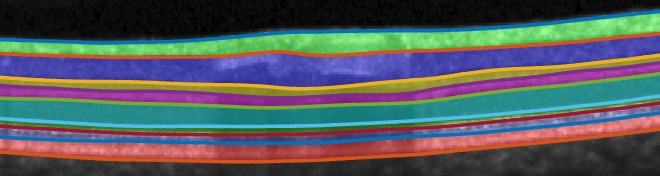}
\end{tabular}
\caption{Left: S-Net results showing defects. Right: R-Net
with the correct topology.}
\label{UN_DNS}
\end{figure}

\begin{figure}[tb!]
\centering
\begin{tabular}{cc}
\includegraphics[width =0.5\textwidth]{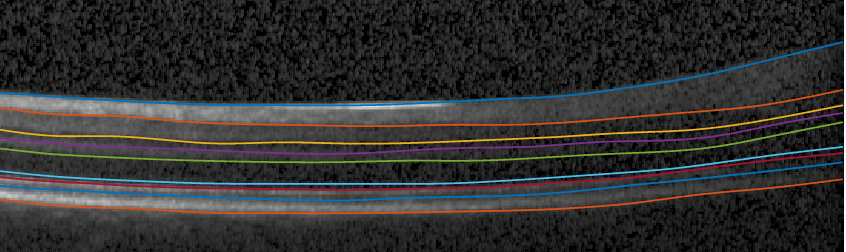}&
\includegraphics[width =0.5\textwidth]{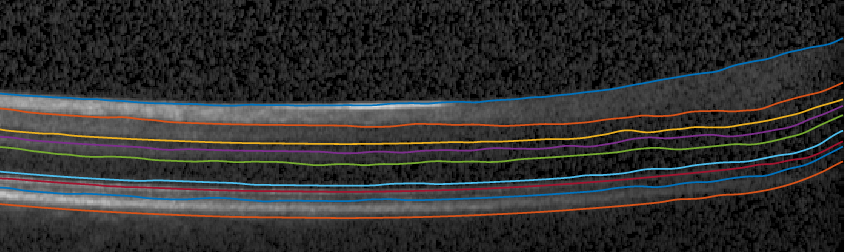}\\
\textbf{(a)} & \textbf{(b)} \\
\includegraphics[width =0.5\textwidth]{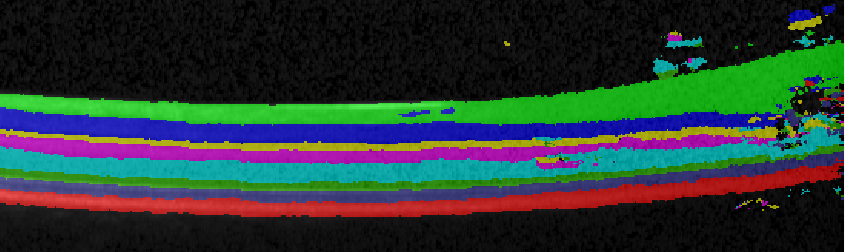}&
\includegraphics[width =0.5\textwidth]{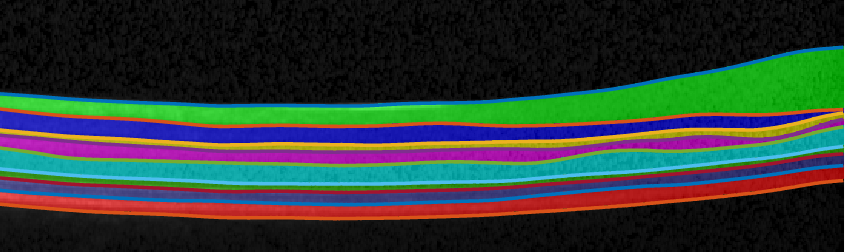}\\
\textbf{(c)} & \textbf{(d)}
\end{tabular}

\caption{\textbf{(a)}~Manual delineation, \textbf{(b)}~RF+G,
\textbf{(c)}~S-Net output, and \textbf{(d)}~R-Net results.}
\label{UN_DNS_DARK}
\end{figure}

Boundary segmentation accuracy was evaluated by comparing the
automatic segmentation results with manual delineation along every
A-scan. The mean absolute distance~(MAD), root mean square error~(RMSE),
and mean signed difference~(MSD) were calculated for the state-of-the-art
RF + Graph method~(RF+G)~\cite{lang2013boe}\footnote{\tiny{Code for
Lang et al. downloaded from
\texttt{https://www.nitrc.org/projects/aura\_tools/}}} and our
proposed deep networks~(S-Net + R-Net). The Wilcoxon signed test was
used to compare these two methods and the 95$\%$ quantile of the MSD
is also reported. These results are shown in Table.~\ref{b_acc}. The depth resolution is 3.9~$\mu$m. From the table, both methods have
MAE and RMSE less than 1 pixel and our proposed method achieves
similar or slightly better results than the state-of-the-art graph
methods. The MSD and 95$\%$ quantile show that compared to our
proposed method, the graph method is more biased. Figs.~\ref{UN_DNS}
and~\ref{UN_DNS_DARK} show some examples that when the image is of
poor quality or the boundaries in the image are not clear, the S-Net
results can be wrong whereas R-Net guarantees the correct topology
while maintaining state-of-the-art accuracy.

The total segmentation time of our proposed deep network for one
$496\times 1024\times 49$ scan is 10 s (preprocessing and
reconstruction included), of which the deep network inference takes
5.85 s. The segmentation is performed with Python~3.6 and the
preprocessing is performed in Matlab R2016b called directly from the
Python environment. The RF+G method, had a total segmentation
time of 100s in Matlab R2016b, of which RF classification was 62 s and
the graph method took 20 s.

\begin{table}[h!]
\centering
\caption{Boundary accuracy evaluated on 20 fully delineated scans for
RF+G~\cite{lang2013boe} and our proposed method, S-Net followed by
R-Net~(S+R-Net).(MAD -- mean absolute distance; RMSE -- root mean
square error; MSD -- mean signed difference; p -- $p$-value.)}
\label{b_acc}
\begin{tabular}{@{}l ccccc c ccccc@{}}
\toprule
& \multicolumn{5}{c}{\textbf{MAD ($\bm{\mu m}$)}} &\hn\hn\hn\hn&
\multicolumn{5}{c}{\textbf{RMSE ($\bm{\mu m}$)}}\\
\cmidrule{2-6}
\cmidrule{8-12}
\textbf{Boundary} & \textbf{RF+G} && \textbf{S+R-Net} && p &&
\textbf{RF+G} && \textbf{S+R-Net} && p\\
\cmidrule{1-12}
\textbf{Vitre-RNFL} & 2.24 && 2.22 && 0.92 && 2.88 && 2.81 && 0.78\\
\textbf{RNFL-GCL} & 2.90 && 2.95 && 0.76 && 4.28 && 4.45 && 0.51\\
\textbf{IPL-INL} & 3.10 && 2.99 && 0.82 && 4.42 && 3.97 && 0.16\\
\textbf{INL-OPL} & 3.09 && 3.22 && 0.20 && 4.31 && 4.18 && 0.90\\
\textbf{OPL-ONL} & 2.74 && 2.78 && 0.76 && 3.92 && 3.82 && 0.95\\
\textbf{ELM} & 2.32 && 2.63 && 0.07 && 2.94 && 3.29 && 0.06\\
\textbf{IS-OS} & 2.38 && 2.12 && 0.76 && 2.91 && 2.62 && 0.97\\
\textbf{OS-RPE} & 3.34 && 3.44 && 0.74 && 4.43 && 4.39 && 0.92\\
\textbf{RPE} & 3.33 && 3.02 && 0.84 && 3.94 && 3.72 && 0.90\\
\cmidrule{1-12}
\textbf{Overall} & 2.83 && 2.82 && 0.42 && 3.78 && 3.69 && 0.56\\
\bottomrule
\end{tabular}
\begin{tabular}{@{}l c c c@{}}
\\[-0.7em]
\toprule
& \multicolumn{3}{c}{\textbf{MSD~($\bm{\mu m}$)~~(95\% quantile)}}\\
\cmidrule{2-4}
\textbf{Boundary} & \textbf{RF+G} &~~~~& \textbf{S+R-Net} \\
\cmidrule{1-4}
\textbf{Vitre-RNFL} & \hn1.02~~(-3.92, \hn6.77) && \hn0.32~~(-5.06,
\hn6.12) \\
\textbf{RNFL-GCL} &-0.37~~(-9.14, \hn7.26) && -0.43~~(-9.13, \hn7.21)
\\
\textbf{IPL-INL} & \hn0.68~~(-8.26, \hn8.70) && \hn0.67~~(-7.03,
\hn8.75) \\
\textbf{INL-OPL} &-0.04~~(-8.70, \hn8.00) && -1.08~~(-9.44, \hn6.89) \\
\textbf{OPL-ONL} & \hn0.53~~(-7.75, \hn7.98) && \hn0.80~~(-6.78,
\hn8.42) \\
\textbf{ELM} & \hn0.13~~(-5.69, \hn6.34) && -0.94~~(-7.51, \hn5.66) \\
\textbf{IS-OS} & \hn1.53~~(-3.09, \hn7.31) && -0.07~~(-4.70, \hn5.83)
\\
\textbf{OS-RPE} & \hn1.15~~(-6.85, 12.20) && -0.06~~(-9.26, 10.39) \\
\textbf{RPE} & \hn2.53~~(-3.27, 12.27) && \hn1.10~~(-5.67, 10.72) \\
\cmidrule{1-4}
\textbf{Overall} & \hn0.80~~(-6.62, \hn9.08) && \hn0.03~~(-7.54,
\hn8.17) \\
\bottomrule
\end{tabular}
\end{table}

\section{Discussion and conclusion}
In this paper, we presented a fast topology guaranteed deep learning
method for retinal OCT segmentation. Our method adds a thickness
regression network after a conventional pixel-wise labeling network
and utilizes the Relu activation to
guarantee the non-negativity of the output and thus guarantee the
topology. Since the R-Net is trained on masks that can be easily
generated, our proposed framework can provide a topology guaranteed
segmentation solution for other layered structures.

%
%
\section{Acknowledgments}

This work was supported by the NIH/NEI under grant R01-EY024655.

\bibliographystyle{splncs03}
\bibliography{ml,oct}

\begin{thebibliography}{10}
\providecommand{\url}[1]{\texttt{#1}}
\providecommand{\urlprefix}{URL }

\bibitem{antony2014miccai}
Antony, B.J., Miri, M.S., Abr\`{a}moff, M.D., Kwon, Y.H., Garvin, M.K.:
  {Automated 3D segmentation of multiple surfaces with a shared hole:
  segmentation of the neural canal opening in SD-OCT volumes}. In:
  17$^{\mbox{\tiny{th}}}$ International Conference on Medical Image Computing
  and Computer Assisted Intervention~(MICCAI~2014). Lecture Notes in Computer
  Science, vol. 8673, pp. 739--746. Springer Berlin Heidelberg (2014)

\bibitem{bentaieb2016miccai}
BenTaieb, A., Hamarneh, G.: {Topology Aware Fully Convolutional Networks for
  Histology Gland Segmentation}. In: 19$^{\mbox{\tiny{th}}}$ International
  Conference on Medical Image Computing and Computer Assisted
  Intervention~(MICCAI~2016). Lecture Notes in Computer Science, vol. 9901, pp.
  460--468. Springer Berlin Heidelberg (2016)

\bibitem{carass2014boe}
Carass, A., Lang, A., Hauser, M., Calabresi, P.A., Ying, H.S., Prince, J.L.:
  {Multiple-object geometric deformable model for segmentation of macular OCT}.
  Biomed. Opt. Express  5(4),  1062--1074 (2014)

\bibitem{dahl2013improving}
Dahl, G.E., Sainath, T.N., Hinton, G.E.: Improving deep neural networks for
  lvcsr using rectified linear units and dropout. In: Acoustics, Speech and
  Signal Processing (ICASSP), 2013 IEEE International Conference on. pp.
  8609--8613. IEEE (2013)

\bibitem{dou20173d}
Dou, Q., Yu, L., Chen, H., Jin, Y., Yang, X., Qin, J., Heng, P.A.: 3d deeply
  supervised network for automated segmentation of volumetric medical images.
  Medical Image Analysis  (2017)

\bibitem{garvin2009tmi}
Garvin, M.K., Abr\`amoff, M.D., Wu, X., Russell, S.R., Burns, T.L., Sonka, M.:
  {Automated 3-D intraretinal layer segmentation of macular spectral-domain
  optical coherence tomography images}. IEEE Trans. Med. Imag.  28(9),
  1436--1447 (2009)

\bibitem{he2017towards}
He, Y., Carass, A., Yun, Y., Zhao, C., Jedynak, B.M., Solomon, S.D., Saidha,
  S., Calabresi, P.A., Prince, J.L.: Towards topological correct segmentation
  of macular oct from cascaded fcns. In: Fetal, Infant and Ophthalmic Medical
  Image Analysis, pp. 202--209. Springer (2017)

\bibitem{hee1995archo}
Hee, M.R., Izatt, J.A., Swanson, E.A., Huang, D., Schuman, J.S., Lin, C.P.,
  Puliafito, C.A., Fujimoto, J.G.: {Optical coherence tomography of the human
  retina}. Arch. Ophthalmol.  113(3),  325--332 (1995)

\bibitem{lang2013boe}
Lang, A., Carass, A., Hauser, M., Sotirchos, E.S., Calabresi, P.A., Ying, H.S.,
  Prince, J.L.: {Retinal layer segmentation of macular OCT images using
  boundary classification}. Biomed. Opt. Express  4(7),  1133--1152 (2013)

\bibitem{long2015cvpr}
Long, J., Shelhamer, E., Darrell, T.: Fully convolutional networks for semantic
  segmentation. In: {The IEEE Conference on Computer Vision and Pattern
  Recognition (CVPR)}. pp. 3431--3440 (June 2015)

\bibitem{medeiros2009iovs}
Medeiros, F.A., Zangwill, L.M., Alencar, L.M., Bowd, C., Sample, P.A., Jr.,
  R.S., Weinreb, R.N.: {Detection of Glaucoma Progression with Stratus OCT
  Retinal Nerve Fiber Layer, Optic Nerve Head, and Macular Thickness
  Measurements}. Invest. Ophthalmol. Vis. Sci.  50(12),  5741--5748 (2009)

\bibitem{rathke2014mia}
Rathke, F., Schmidt, S., Schn\"{o}rr, C.: {Probabilistic Intra-Retinal Layer
  Segmentation in 3-D OCT Images Using Global Shape Regularization}. Medical
  Image Analysis  18(5),  781--794 (2014)

\bibitem{romero2017image}
Romero, A., Drozdzal, M., Erraqabi, A., J{\'e}gou, S., Bengio, Y.: Image
  segmentation by iterative inference from conditional score estimation. arXiv
  preprint arXiv:1705.07450  (2017)

\bibitem{ronneberger2015miccai}
Ronneberger, O., Fischer, P., Brox, T.: {U-Net: Convolutional Networks for
  Biomedical Image Segmentation}. In: 18$^{\mbox{\tiny{th}}}$ International
  Conference on Medical Image Computing and Computer Assisted
  Intervention~(MICCAI~2015). Lecture Notes in Computer Science, vol. 9351, pp.
  234--241. Springer Berlin Heidelberg (2015)

\bibitem{roy2017arxiv}
Roy, A.G., Conjeti, S., Karri, S.P.K., Sheet, D., Katouzian, A., Wachinger, C.,
  Navab, N.: {ReLayNet: Retinal Layer and Fluid Segmentation of Macular Optical
  Coherence Tomography using Fully Convolutional Network}. CoRR  abs/1704.02161
  (2017)

\bibitem{Zheng_2015_ICCV}
Zheng, S., Jayasumana, S., Romera-Paredes, B., Vineet, V., Su, Z., Du, D.,
  Huang, C., Torr, P.H.S.: Conditional random fields as recurrent neural
  networks. In: The IEEE International Conference on Computer Vision (ICCV)
  (December 2015)

\end{thebibliography}

\end{document}